\pdfoutput=1

\documentclass{article}

\usepackage[preprint]{wenxiaobai}

\usepackage[utf8]{inputenc}
\usepackage[T1]{fontenc}
\usepackage{hyperref}
\usepackage{url}
\usepackage{booktabs}
\usepackage{amsfonts}
\usepackage{amsmath}
\usepackage{amssymb}
\usepackage{microtype}
\usepackage{inconsolata}
\usepackage{xcolor}

\usepackage{graphicx}
\usepackage{multirow}
\usepackage{subfigure}
\usepackage{overpic}
\usepackage{tabularx}
\usepackage{wrapfig}
\usepackage{tcolorbox}
\tcbuselibrary{skins,breakable}
\graphicspath{{picture/}}

\newtcolorbox{promptbox}[1][]{
  colback=white,
  colframe=gray!75!black,
  colbacktitle={rgb,255:red,248;green,244;blue,239},
  coltitle=black,
  title=#1,
  fonttitle=\bfseries\small,
  fontupper=\small\ttfamily,
  boxrule=0.8pt,
  arc=2pt,
  left=2pt, right=2pt, top=2pt, bottom=2pt
}

\definecolor{myblue}{RGB}{31,119,180}
\definecolor{mygreen}{RGB}{44,160,44}
\definecolor{myred}{RGB}{214,39,40}

\title{Wiki Live Challenge: Challenging Deep Research Agents with Expert-Level Wikipedia Articles}

\author{
  \bf Shaohan Wang$^{1*}$, Benfeng Xu$^{1,2\dagger}$, Licheng Zhang$^{1\S}$, Mingxuan Du$^{1}$, Chiwei Zhu$^{1}$,\\\bf Xiaorui Wang$^{2}$, Zhendong Mao$^{1}$ and Yongdong Zhang$^{1}$ \\[0.5em]
  $^{1}$University of Science and Technology of China, Hefei, China \\
  $^{2}$Metastone Technology, Beijing, China \\[0.3em]
  \texttt{\{wsh2000, benfeng, zlczlc\}@mail.ustc.edu.cn, zdmao@ustc.edu.cn}
}

\begin{document}
\makeatletter
\renewcommand{\@noticestring}{\rule{0.4\linewidth}{0.4pt}\\[0.5em]$^*$Work done during the internship at Metastone.\\$^\dagger$Project lead. $^\S$Corresponding author.\\Preprint. Work in progress.}
\makeatother
\maketitle

\begin{abstract}
\hspace{2em}Deep Research Agents (DRAs) have demonstrated remarkable capabilities in autonomous information retrieval and report generation, showing great potential to assist humans in complex research tasks. Current evaluation frameworks primarily rely on LLM-generated references or LLM-derived evaluation dimensions. While these approaches offer scalability, they often lack the reliability of expert-verified content and struggle to provide objective, fine-grained assessments of critical dimensions. To bridge this gap, we introduce \textbf{Wiki Live Challenge (WLC)}, a live benchmark that leverages the newest Wikipedia Good Articles (GAs) as expert-level references. Wikipedia's strict standards for neutrality, comprehensiveness, and verifiability serve as a great challenge for DRAs, with GAs representing the pinnacle of which. We curate a dataset of 100 recent Good Articles and propose \textbf{Wiki Eval}, a comprehensive evaluation framework comprising a fine-grained evaluation method with 39 criteria for writing quality and rigorous metrics for factual verifiability. Extensive experiments on various DRA systems demonstrate a significant gap between current DRAs and human expert-level Wikipedia articles, validating the effectiveness of WLC in advancing agent research. We release our benchmark at \url{https://github.com/WangShao2000/Wiki_Live_Challenge}
\end{abstract}

\section{Introduction}

\begin{wrapfigure}[22]{r}{0.5\linewidth}
  \centering
  \vspace{-14pt}
  \includegraphics[width=\linewidth]{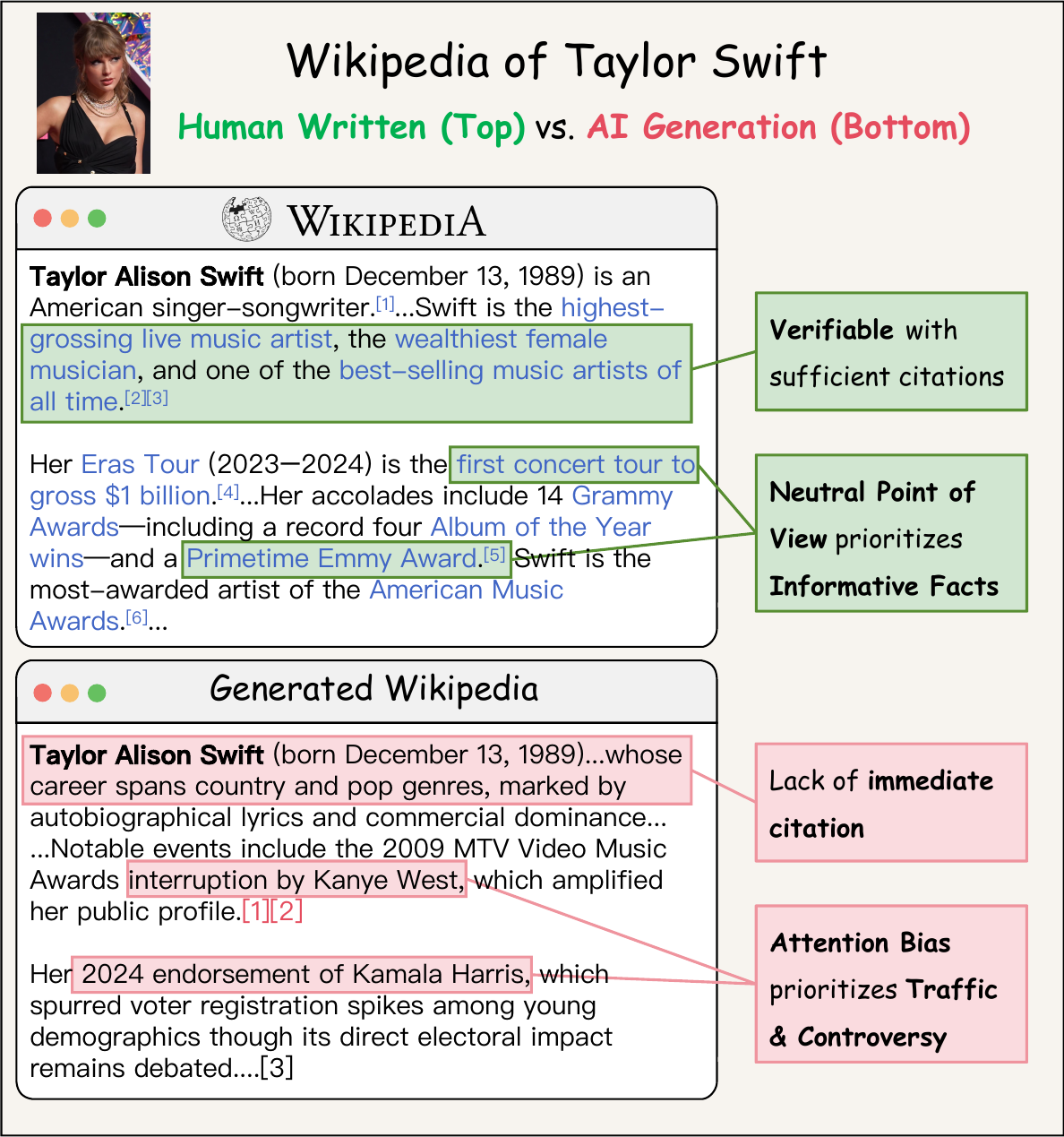}
  \caption{\textbf{Gap between human-written and AI-generated Wikipedia articles.} Human-authored articles (top) feature rigorous citations and neutral tone, while AI-generated ones (bottom) lack citations and exhibit bias toward trending topics.}
  \label{fig:moti}
  \vspace{-10pt}
\end{wrapfigure}

With the explosive growth of Large Language Model (LLM) capabilities, LLM-driven agents have demonstrated remarkable potential in handling expert-level tasks~\citep{achiam2023gpt, yang2025qwen3, team2025kimi, mialon2024gaia}. These agents are capable of multi-step task planning, tool utilization, and interaction with real-world environments to accomplish complex objectives.
Among these, the Deep Research Agent (DRA) represents one of the most advanced agent systems~\citep{team2025tongyi, qiao2025webresearcherunleashingunboundedreasoning, li2025websailornavigatingsuperhumanreasoning, zheng-etal-2025-deepresearcher}. By performing multi-step web information retrieval, integration, and reasoning, DRAs can complete research tasks that would typically require significant time and effort from human experts.

However, existing DRAs still suffer from issues such as hallucinations and biases in research and writing. To comprehensively evaluate the capabilities of these systems, two core challenges must be addressed: how to efficiently obtain reliable expert-level articles as references, and how to design an objective evaluation method that comprehensively reflects DRA research capability and writing quality~\citep{du2025deepresearchbenchcomprehensivebenchmark, xu2025researcherbenchevaluatingdeepai}.

Existing efforts~\citep{du2025deepresearchbenchcomprehensivebenchmark, li2025reportbenchevaluatingdeepresearch} have attempted to address these challenges by using reports generated by strong DRAs as references to evaluate others, scoring them based on manually designed criteria such as comprehensiveness, depth and so on. While these methods can reflect the gap between models to some extent, the reference reports are LLM-generated and lack quality assurance. Furthermore, evaluation criteria are often directly defined by LLMs or rely on internal model knowledge for verification, which may lead to results that deviate from human expert expectations~\citep{fan2025understandingdeepresearchreports, li2024llmsasjudgescomprehensivesurveyllmbased}. Other approaches attempt to design specific rubrics for different reports~\citep{sharma2025researchrubricsbenchmarkpromptsrubrics, xu2025researcherbenchevaluatingdeepai}; however, these rubrics are often coarse-grained generated by LLMs or require additional human annotation.

To bridge this gap, we introduce the \textbf{Wiki Live Challenge (WLC)}, a live benchmark designed to challenge DRAs with the latest Wikipedia Good Articles. Wikipedia articles represent comprehensive research on a subject, involving extensive information gathering and organization, written from an objective and neutral perspective with strict verifiability. We posit that the ability to produce such content comprehensively reflects a DRA's capabilities, yet current models fall considerably short of these standards, as illustrated in Figure~\ref{fig:moti}. Therefore, we utilize high-quality Wikipedia Good Articles as real-time human expert references and, based on their corresponding criteria, assess the disparity between DRA-generated reports and real-world human-authored content.

Specifically, we collected 100 expert-level Wikipedia Good Articles that strictly follow Wikipedia's editorial guidelines and have been reviewed and revised by human experts, providing strong quality assurance. We ensure that the article set is recent and will be continuously updated to avoid data contamination. Leveraging these articles as human-expert references, we construct \textbf{Wiki Eval}, an evaluation framework grounded in Wikipedia Good Article criteria. This framework comprises two key components: \textbf{Wiki Writing} and \textbf{Wiki Fact}. Wiki Writing serves as a fine-grained writing evaluation protocol with 39 criteria covering GA-aligned writing dimensions. Additionally, Wiki Fact assesses the DRA's information retrieval capability and factual reliability following the verifiability criteria, with two sub-metrics that measure (i) information richness relative to Wikipedia and (ii) whether generated statements are strictly traceable to supporting sources.

Our contributions are summarized as follows:
\begin{itemize}
\item We introduce \textbf{Wiki Live Challenge (WLC)}, a live benchmark designed to challenge the capability of DRAs in writing Wikipedia. Sourced from Wikipedia Good Articles, our benchmark ensures high quality through expert review and validation, serving as a reliable human expert reference.
\item We propose \textbf{Wiki Eval}, an evaluation framework focusing on both writing quality and factuality, with all criteria strictly grounded in Wikipedia Good Article criteria.
\item We conducted extensive experiments across a diverse set of DRA systems and performed comprehensive analyses and human studies to validate the reliability of our evaluation framework. We also plan to maintain and extend the benchmark to better reflect evolving real-world conditions.
\end{itemize}

\begin{figure}[!htbp]
  \centering
  \includegraphics[width=0.95\linewidth]{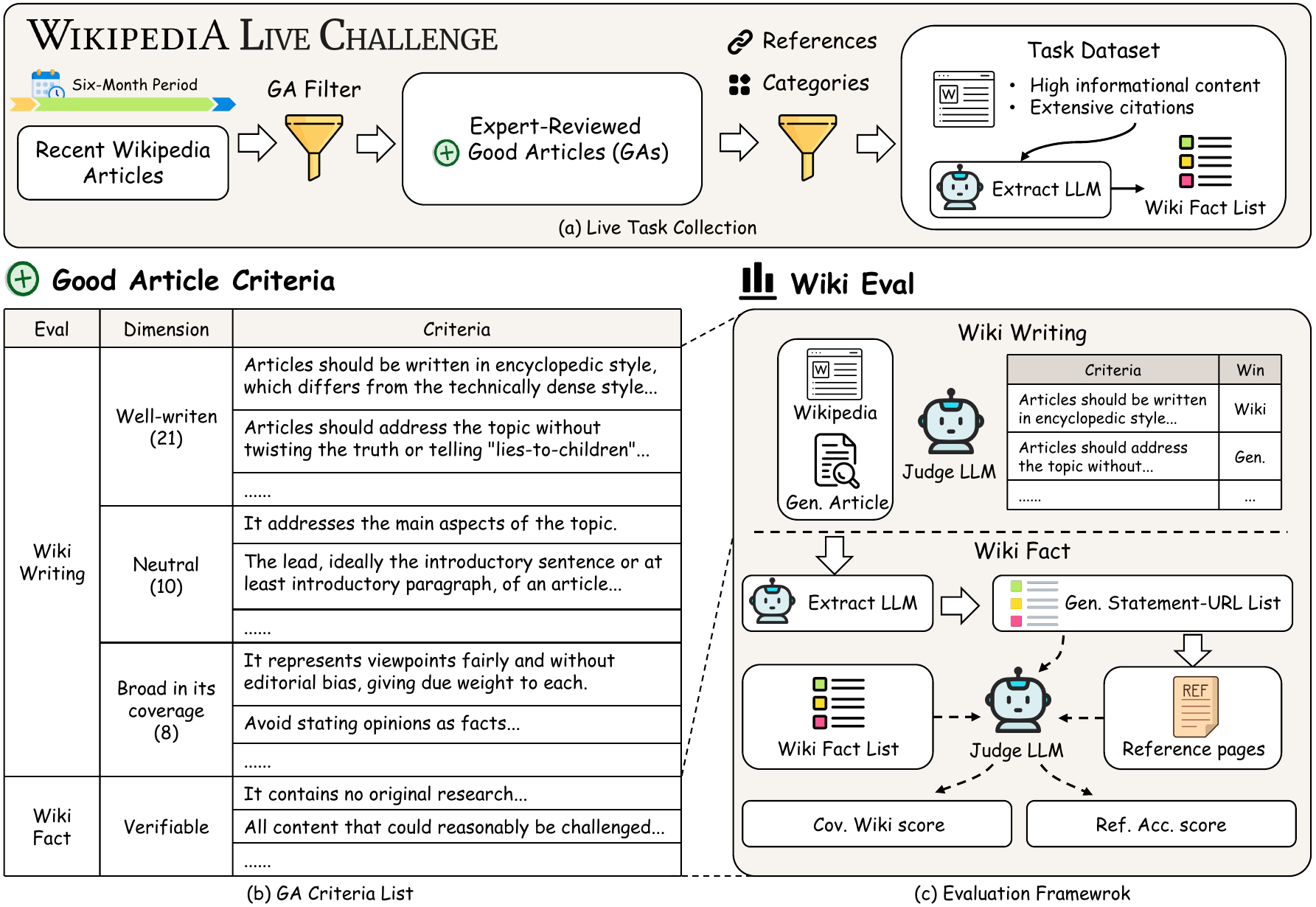}
  \caption{\textbf{The overview of our Wiki Live Challenge (WLC) benchmark.} (a) We continuously collect recent Wikipedia articles (e.g., from Mar.\ 1 to Dec.\ 1 in this iteration), filter the latest expert-reviewed Good Articles, and build the live task dataset.\ (b) We strictly grounded in GA criteria: well-written, neutral, broad coverage and verifiable.\ (c) Our evaluation framework, Wiki Eval, incorporates two key dimensions: Wiki Writing and Wiki Fact.}
  \label{fig:main}
\end{figure}

\section{Related Work}

\subsection{Deep Research Agent}
Deep Research Agents (DRAs) are designed to autonomously explore the web, retrieve information, and synthesize findings into comprehensive reports. Recent progress focuses on improving their reasoning and planning for long-term tasks. Notably, DeepResearcher~\citep{zheng-etal-2025-deepresearcher} is the first work to train LLMs via end-to-end reinforcement learning in a real, dynamic web environment for deep information retrieval and integration. Similarly, Tongyi DeepResearch~\citep{team2025tongyi} employs an end-to-end training framework to enable scalable reasoning. WebResearcher~\citep{qiao2025webresearcherunleashingunboundedreasoning} treats research as a decision process, using iterative refinement to manage noise. WebSailor~\citep{li2025websailornavigatingsuperhumanreasoning} tackles uncertainty in web navigation through structured sampling. Furthermore, systems like WebDancer~\citep{wu2025webdancerautonomousinformationseeking} have pushed the boundaries of autonomous information seeking. These developments highlight a shift towards agents that can independently verify information and synthesize knowledge.

\subsection{Deep Research Benchmarks}
Evaluating the capabilities of DRAs requires benchmarks that go beyond simple question answering. Early general agent benchmarks such as GAIA~\citep{mialon2024gaia} and AgentBench~\citep{liu2025agentbenchevaluatingllmsagents} assess fundamental abilities like tool use and reasoning but often lack the depth required for evaluating long-form research reports. FreshWiki~\citep{shao-etal-2024-assisting} is an early dataset utilizing Wikipedia articles to evaluate generated text, but it lacks fine-grained rubrics tailored for modern DRAs. DeepResearch Bench~\citep{du2025deepresearchbenchcomprehensivebenchmark} provides 100 PhD-level tasks across 22 domains, employing reference-based adaptive criteria. ReportBench~\citep{li2025reportbenchevaluatingdeepresearch} focuses on report generation quality using survey papers as references, while ResearchRubrics~\citep{sharma2025researchrubricsbenchmarkpromptsrubrics} offers expert-written criteria for evaluating open-ended queries. More recent benchmarks further emphasize live tasks and expert-grounded evaluation, including LiveResearchBench~\citep{wang2025liveresearchbenchlivebenchmarkusercentric}, DeepScholar-Bench~\citep{patel2025deepscholarbenchlivebenchmarkautomated}, and DEER~\citep{han2025deercomprehensivereliablebenchmark}. Additionally, BrowseComp~\citep{wei2025browsecompsimplechallengingbenchmark}, and CiteEval~\citep{xu2025citeevalprincipledrivencitationevaluation} focus on citation accuracy and source attribution. Despite these efforts, existing benchmarks often rely on static or model-generated references lacking rigorous human verification, which limits their ability to objectively measure alignment with expert standards in real-world scenarios.

\begin{figure}[!htbp]
  \centering
  \includegraphics[width=0.9\linewidth]{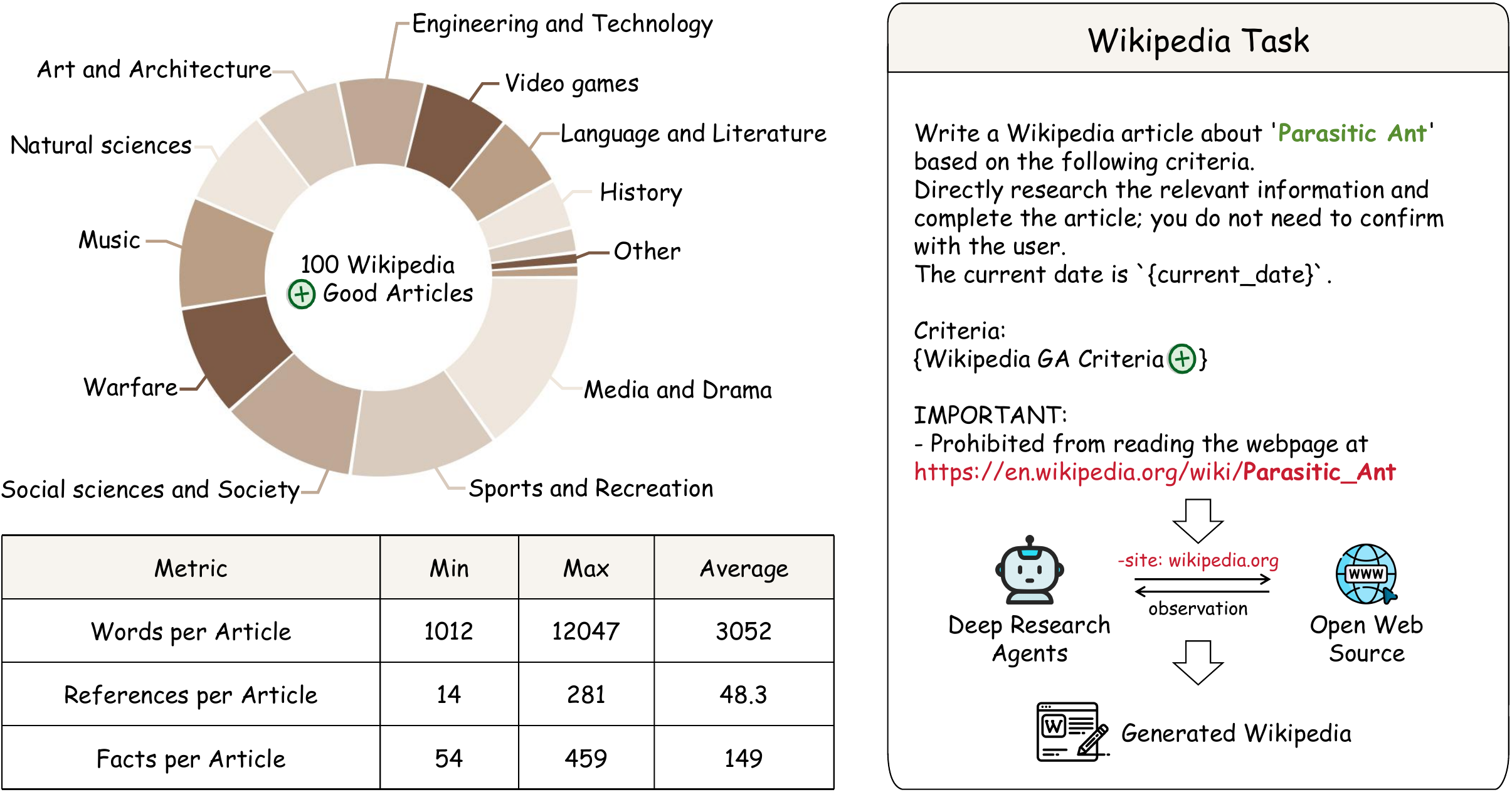}
  \caption{\textbf{Overview of the WLC benchmark dataset.} The left panel displays the distribution of collected Wikipedia Good Articles across 15 major categories and the key statistics of the WLC Benchmark Dataset. The right panel illustrates a representative task case.}
  \label{fig:task}
\end{figure}

\section{Wiki Live Challenge}

In this section, we provide a detailed introduction to the Wiki Live Challenge (WLC). We first outline our data collection and construction methodology, followed by our Wiki Eval framework based on the Wikipedia Good Article criteria. Overall framework is illustrated in Figure~\ref{fig:main}.

\subsection{Data Construction}\label{sec:3.1}

Existing benchmarks for Deep Research Agents typically rely on reports generated by strong models or collected from the open web as references~\citep{du2025deepresearchbenchcomprehensivebenchmark, wang2025liveresearchbenchlivebenchmarkusercentric, patel2025deepscholarbenchlivebenchmarkautomated, han2025deercomprehensivereliablebenchmark}. However, these sources often lack expert evaluation, offering no guarantees of quality. Furthermore, they may contain biases or errors stemming from the inherent biases of the generative models or human authors. In contrast, Wikipedia serves as a reliable knowledge source rigorously reviewed and revised by human editors, adhering to strict core policies that ensure neutrality, comprehensiveness, and strict verifiability. Particularly, Good Articles (GAs) represent the pinnacle of this quality, undergoing a meticulous expert review process to meet the highest editorial standards. We posit that the ability to autonomously author such content---necessitating extensive research, the synthesis of diverse viewpoints, and the maintenance of objectivity---serves as a rigorous challenge for DRAs. Therefore, we utilize these high-quality GAs as reference standards and construct our evaluation methodology grounded in the official Good Article criteria.

Specifically, we collected all new Wikipedia articles created between March 1, 2025, and December 1, 2025\footnote{This timeframe utilizes the most recent Wikipedia articles to mitigate potential data contamination, ensuring the content postdates the knowledge cutoff of current mainstream models.}. From this corpus, we filtered for Good Articles (GA), identifying 304 articles that have passed Wikipedia's review process and strictly adhere to the Good Article criteria, thus ensuring high quality. To ensure the complexity of the task, we ranked these articles based on the number of reference URLs and structural depth, excluding simple list-based articles. Ultimately, we curated a dataset of 100 Wikipedia Good Articles to serve as our benchmark. The distribution of these articles, along with the dataset statistics and a representative task case, is shown in Figure~\ref{fig:task}, which covers 15 major domains, providing a comprehensive benchmark for Deep Research Agents.

\subsection{Evaluation Methodology}\label{sec:3.2}

Our evaluation framework is grounded in the Wikipedia Good Article criteria, a widely recognized standard established by human experts\footnote{\url{https://en.wikipedia.org/wiki/Wikipedia:Good_article_criteria}}. We selected four dimensions most relevant to DRA capabilities, focusing on writing style, neutral point of view, broad in coverage, and verifiability. Based on these, we constructed \textbf{Wiki Eval}, an evaluation methodology comprising two primary components: Wiki Writing and Wiki Fact.

\subsubsection{Wiki Writing}

To assess writing quality, we construct Wiki Writing, a fine-grained evaluation framework based on the three core dimensions of the Wikipedia Good Article criteria: \textit{Well-written}, \textit{Neutral}, and \textit{Broad in its coverage}. This framework comprises 39 distinct criteria derived directly from official Wikipedia writing guidelines, as shown in Figure~\ref{fig:main}-\textit{b}. Subsequently, we employ an LLM-as-a-Judge approach. For each criterion, we provide the evaluation model with both the original Wikipedia article and the LLM-generated article to determine the winner for this criterion:
\begin{equation}
  \label{eq:judge}
  \text{Judge}(w_i, g_i) = \text{Judge-LLM}(w_i, g_i).
\end{equation}
where $w_i$ denotes the Wikipedia reference, $g_i$ represents the generated article, and $\text{Judge-LLM}$ is the model used to determine the superior output for the given criterion. Finally, we calculate the score for each criterion based on the model's judgment and aggregate these scores to compute the overall writing score for the article. Detailed information regarding the criteria sources and the complete list of criteria are provided in the Appendix~\ref{sec:appendixC}.

\subsubsection{Wiki Fact}

To evaluate the article's factual accuracy (1) relative to Wikipedia and (2) relative to cited references, we introduce two evaluation metrics, as illustrated in Figure~\ref{fig:main}-\textit{c}. We first preprocess both the Wikipedia article and the generated article by employing an extraction LLM to extract facts from each, yielding a fact list for the Wikipedia article and a Statement-URL pair list for the generated article.

For factual accuracy \textbf{relative to Wikipedia}, we measure the coverage of generated statements against Wikipedia facts. For each fact $f_i$ in Wikipedia fact list, we retrieve the top-10 most relevant statements from the generated article. These statements and target fact are then input into a fact-checking model to determine a consistency score:
\begin{equation}
  \label{eq:fact}
  \text{Fact}(f_i, G) = 
  \begin{cases}
    1, & \text{if consistent} \\
    0, & \text{if inconsistent} \\
    0, & \text{if conflict}
  \end{cases}
\end{equation}

We then calculate the coverage score for each article by averaging the scores of all its target facts:
\begin{equation}
  \label{eq:coverwiki}
  \text{Cov. Wiki.} = \frac{1}{|F|} \sum_{f_i \in F} \text{Fact}(f_i, G)
\end{equation}
where $F$ represents the set of facts extracted from the Wikipedia article, and $G$ represents the set of facts from the corresponding generated article.

For factual accuracy \textbf{relative to references}, we assess whether the generated statements are supported by their citations. For each extracted Statement-URL pair, we utilize Jina Reader\footnote{\url{https://jina.ai}} to retrieve the content of the cited webpage. We then employ the fact-checking model to verify if the statement is supported by source content.

The final Reference Accuracy score is calculated as the proportion of statements that are fully supported by their references set $R$:
\begin{equation}
  \label{eq:refacc}
  \text{Ref. Acc.} = \frac{1}{|S|} \sum_{s_i \in S} \text{Fact}(s_i, R)
\end{equation}
where $S$ denotes the list of Statement-URL pairs in the generated article.

\section{Experimental Settings}

\subsection{Implementation Details}

For the evaluation of Wiki Writing, we utilize Gemini-2.5-pro~\citep{comanici2025gemini} as the Judge LLM.\ For the Wiki Fact assessment, we employ Gemini-2.5-flash as both the extraction and fact-checking LLM, balancing performance with cost-effectiveness given the high token consumption. All reference Wikipedia pages were collected on December 15, 2025, and all results presented in Table~\ref{main-result} are based on evaluations against these 100 complete Wikipedia articles. Further implementation settings are provided in Appendix~\ref{sec:appendixA.1}.

\subsection{Evaluated Models}

In this work, we extensively evaluated advanced Deep Research Agent systems, including proprietary systems such as OpenAI o3 Deep Research~\citep{OpenAIDeepResearch}, Gemini-2.5-pro Deep Research~\citep{GeminiDeepResearch}, and Qwen-3-max Deep Research~\citep{yang2025qwen3}. Regarding open-source frameworks, we selected Tongyi DeepResearch~\citep{team2025tongyi} and Deep Researcher~\citep{zheng-etal-2025-deepresearcher}, two state-of-the-art open-source models trained via reinforcement learning. Additionally, we evaluated LangChain Open Deep Research\footnote{\url{https://github.com/langchain-ai/open_deep_research}}, an open-source framework powered by proprietary models, utilizing GPT-4.1 and GPT-5 as backends. All model articles were collected between December 15 and 19, 2025.

\begin{table}[!htbp]
  \centering
  \caption{\textbf{Main results of WLC across Wiki Writing and Wiki Fact.} Wiki Writing are computed by aggregating wins over our 39 Wiki GA-based criteria. Cov. Wiki measures factual coverage against the extracted Wikipedia fact list, and Ref. Acc.\ measures the proportion of cited statements that are supported by their referenced webpages; missing entries indicate that reliable Statement-URL extraction was not possible due to citation formatting.}
  \label{main-result}
  \small
  \renewcommand{\arraystretch}{1.15}
  \begin{tabular*}{\linewidth}{@{\extracolsep{\fill}}lcccccc}
    \toprule
    \multirow{2}{*}{\textbf{Models}} & \multicolumn{4}{c}{\textbf{Wiki Writing}} & \multicolumn{2}{c}{\textbf{Wiki Fact}}\\
    \cmidrule(lr){2-5} \cmidrule(lr){6-7}
    & Overall & Well-writ. & Neutral & Broad & Cov. Wiki & Ref. Acc.\\
    \midrule
    \multicolumn{7}{c}{\textit{Open-Source Agent Framework}}\\
    \midrule
    Deep Researcher & 2.28 & 1.90 & 1.90 & 3.75 & 5.62 & --\\
    Tongyi Deep Research & 15.05 & 10.90 & 11.60 & 30.25 & 22.73 & --\\
    Langchain (GPT-4.1) & 20.67 & 18.76 & 19.40 & 27.25 & 7.08 & 7.34\\
    Langchain (GPT-5) & \underline{53.62} & \underline{50.95} & \textbf{54.20} & \underline{59.88} & 20.96 & \textbf{67.60}\\
    \midrule
    \multicolumn{7}{c}{\textit{Proprietary Agent Framework}}\\
    \midrule
    Doubao Deep Research & 19.13 & 16.00 & 16.10 & 31.13 & 22.97 & 37.05\\
    Qwen-3-max Deep Res. & 25.15 & 18.29 & 27.70 & 40.00 & 22.22 & 61.44\\
    Perplexity Deep Res. & 27.38 & 26.79 & 20.40 & 37.63 & \underline{29.21} & 28.27\\
    Grok Deep Search & 28.38 & 27.52 & 24.30 & 35.75 & 20.73 & 60.63\\
    OpenAI o3 Deep Res. & 31.08 & 28.43 & 24.90 & 45.75 & 25.12 & 57.44\\
    Gemini-2.5-pro Deep Res. & 35.18 & 30.10 & 26.10 & \underline{59.88} & \textbf{30.76} & 41.68\\
    Gemini-3-pro Deep Res. & \textbf{58.33} & \textbf{60.81} & \underline{46.10} & \textbf{67.12} & 28.83 & \underline{66.98}\\
    \bottomrule
  \end{tabular*}
\end{table}

\section{Results and Discussions}

\subsection{Main Results}

\subsubsection{Evaluation on Wiki Writing}

As shown in Table~\ref{main-result}, among the various Deep Research Agents, Gemini-3-pro Deep Research and the LangChain framework powered by GPT-5 demonstrates a significant advantage, with Gemini-2.5-pro and OpenAI o3 Deep Research also exhibiting strong performance. We observe substantial performance disparities across different DRAs. Notably, fully open-source DRA frameworks lag significantly behind proprietary models. Deep Researcher, for instance, achieved a score of only 2.28, which is markedly lower than other models. Manual review revealed that its reports are often incomplete, typically concluding after minimal information gathering steps, which adversely affected its scores on writing criteria. Tongyi DeepResearch performed relatively better, achieving scores comparable to some proprietary models like Doubao Deep Research; however, a gap remains compared to state-of-the-art DRA frameworks. This suggests that achieving the long-form report generation capabilities of proprietary models remains a challenge for smaller models trained via end-to-end RL.

\subsubsection{Evaluation on Wiki Fact}

Considering Wiki Fact in Table~\ref{main-result}, we observe that all DRA systems perform poorly in terms of coverage of Wikipedia facts. Even the best-performing agent, Gemini-2.5-pro Deep Research, achieved an average knowledge coverage of only 30.76\%, indicating that current models are still far from reaching the expert-level information gathering capabilities of Wikipedia. Notably, while LangChain (GPT-5) excels in writing, its knowledge retrieval performance lags behind proprietary frameworks.

\begin{figure}[!htbp]
  \centering
  \includegraphics[width=\linewidth]{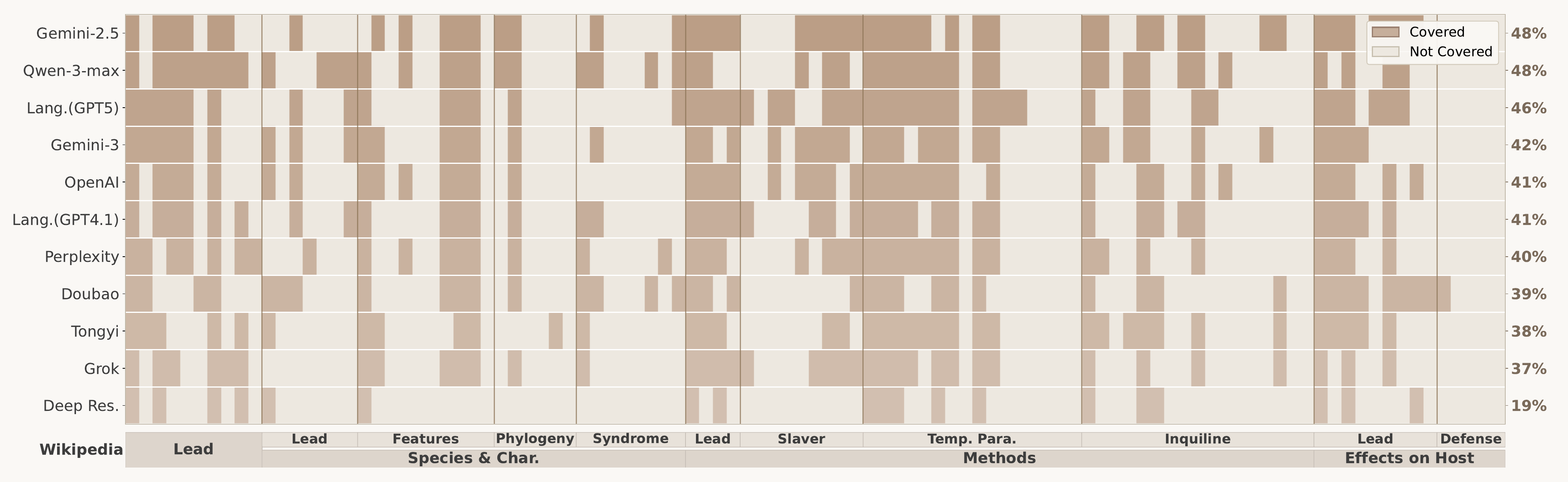}
  \caption{\textbf{Fact Coverage Heatmap on Wikipedia Good Article ``Parasitic Ant''.} The x-axis represents individual facts ordered by their appearance in the article sections, and the y-axis represents different DRAs.}
  \label{fig:heatmap}
\end{figure}

To further investigate this gap, we present a case study on the \textit{Parasitic Ant} article in Figure~\ref{fig:heatmap}. The heatmap reveals a difficulty gradient aligned with article structure: DRAs perform well on procedural sections like \textit{Methods} but struggle with specialized sections such as \textit{Phylogeny} and \textit{Defense}. Analysis shows that while general definitions are universally covered, all systems fail to retrieve precise quantitative data (e.g., gene counts) and domain-specific terminology. This suggests that agents effectively capture broad concepts but lack the precision for granular, high-difficulty details.

Regarding Reference Accuracy, scores for Open-Source models are omitted due to their frequent failure to generate properly formatted citation markers. For LangChain (GPT-4.1), the scarcity of citations led to low scores, whereas GPT-5 demonstrated superior performance with near-complete citation coverage. Proprietary frameworks like Perplexity scored lower likely because they only cite a subset of retrieved content, leaving some statements unverified. Given the lack of transparency in internal search mechanisms and inconsistencies in citation placement or accessibility (e.g., paywalls), this score serves as a supplementary metric reflecting the verifiability of accessible information rather than a definitive measure of grounding.

\begin{table}[!htbp]
  \centering
  \caption{\textbf{Conflict rates across different systems.} Wiki Conf. is the fraction of statements conflicting with Wikipedia facts. Ref. Conf. is the fraction of statements conflicting with their cited references.}
  \label{tab:conflict}
  \small
  \renewcommand{\arraystretch}{1.15}
  \begin{tabular*}{0.8\linewidth}{@{\extracolsep{\fill}}lcc}
    \toprule
    \textbf{Models} & \textbf{Wiki Conf. (\%)} & \textbf{Ref. Conf. (\%)} \\
    \midrule
    Deep Researcher & 9.98 & -- \\
    Tongyi Deep Research & 9.89 & -- \\
    Langchain (GPT-5) & 10.45 & 3.14 \\
    Langchain (GPT-4.1) & 24.69 & 2.94 \\
    \midrule
    Gemini-3-pro Deep Research & 7.62 & 4.23 \\
    Perplexity Deep Research & 7.74 & 3.04 \\
    Gemini-2.5-pro Deep Research & 7.79 & 4.58 \\
    OpenAI o3 Deep Research & 9.12 & 3.95 \\
    Doubao Deep Research & 9.78 & 4.65 \\
    Qwen-3-max Deep Research & 10.41 & 6.87 \\
    Grok Deep Search & 11.40 & 4.96 \\
    \bottomrule
  \end{tabular*}
\end{table}

\subsection{Analysis and Discussion}

\subsubsection{Analysis of Fact Conflicts}\label{sec:conflicts}

We further analyze \textit{conflicts}, where generated statements directly contradict either the established facts in Wikipedia Good Articles (\textbf{Wiki Conf.}) or the content of their own cited references (\textbf{Ref. Conf.}). Such conflicts are particularly detrimental as they introduce explicit falsehoods rather than mere omissions.

Table~\ref{tab:conflict} illustrates distinct error patterns across systems, decoupling the analysis of verification against ground truth versus cited sources. A high citation conflict rate points to severe hallucination, where the model fabricates information not supported by its claimed references; notably, Qwen-3-max Deep Research exhibits a high citation conflict rate of 6.87\%. Conversely, a high conflict rate with Wikipedia implies the inclusion of hallucinations or information from unreliable sources that contradicts established facts. For example, while LangChain (GPT-4.1) achieves the lowest citation conflict (2.94\%), it records the highest Wiki conflict (24.69\%), suggesting it may be incorporating incorrect information despite adhering to its own retrieval context. LangChain (GPT-5) demonstrates superior performance by maintaining low conflict rates across both dimensions.

\subsubsection{Analysis Across Categories}

\begin{table}[!htbp]
  \centering
  \caption{\textbf{Correlation between Wikipedia article features and task difficulty.} Task difficulty shows moderate correlation with popularity (page views) but is independent of article length, statement or link count.}
  \label{tab:correlation}
  \small
  \renewcommand{\arraystretch}{1.2}
  \begin{tabular*}{0.8\linewidth}{@{\extracolsep{\fill}}lcl}
    \toprule
    \textbf{Feature} & \textbf{Coeff.} & \textbf{Strength} \\
    \midrule
    Page Views & $r = +0.482$ & Moderate \\
    External Links & $r = +0.042$ & Negligible \\
    Length (Bytes) & $r = -0.027$ & Negligible \\
    Statement Count & $r = +0.152$ & Negligible \\
    \bottomrule
  \end{tabular*}
\end{table}

\paragraph{Difficulty}
Our benchmark comprises Wikipedia articles from 15 distinct categories, revealing significant performance disparities among models across these domains. Notably, in \textit{History} and \textit{Mathematics}, the average win rate across all systems remained below 20\%, whereas in \textit{Natural Sciences} and \textit{Philosophy and Religion}, average scores exceeded 40\%. This highlights substantial variation in the difficulty of information retrieval and summarization across fields. We further analyzed the correlation between Wikipedia statistical features and task difficulty, as detailed in Table~\ref{tab:correlation}. Results indicate that difficulty has negligible correlation with article length or citation count but is moderately correlated with total page views. This suggests that difficulty is largely determined by the complexity of web-based research: higher view counts typically correspond to more popular categories, where information is more accessible for DRAs to mine. More information are provided in Appendix~\ref{sec:appendixB.1}.

\paragraph{Robustness}
To verify the robustness of evaluation across different Wikipedia categories, we further analyzed the performance variation of DRAs across categories. To control for differences caused by inherent category difficulty, we used the deviation of each DRA system's score from the category mean as its relative score. We hypothesized that there is no significant difference in the relative performance of DRAs evaluated across different categories and conducted an ANOVA test. The results indicate that, with the exception of Deep Researcher, there are no significant differences in the relative performance of the systems evaluated across categories ($p>0.05$), demonstrating that our evaluation possesses cross-category robustness. Deep Researcher consistently underperformed across varying task difficulties; thus, we attribute the variance in its evaluation across categories to intrinsic performance limitations. Further details are provided in Appendix~\ref{sec:appendixB.2}.

\subsubsection{Different Models for Judgement}

\begin{table}[!htbp]
  \centering
  \caption{\textbf{Pairwise Agreement Rate (PAR) of different Judge LLMs with human annotations.} Cost represents the average cost per article evaluation. Qwen3-80B-A3B is deployed locally, incurring no API costs.}
  \label{tab:par_results}
  \small
  \renewcommand{\arraystretch}{1.2}
  \begin{tabular*}{0.8\linewidth}{@{\extracolsep{\fill}}lcc}
    \toprule
    \textbf{Judge Models} & \textbf{PAR (\%)} & \textbf{Cost (\$)} \\
    \midrule
    Gemini-2.5-pro & \textbf{83.59} & 0.132 \\
    GPT-5 & 80.67 & 0.128 \\
    Qwen3-235B-A22B & 78.72 & 0.071 \\
    DeepSeek-V3.1 & 76.41 & \textbf{0.024} \\
    GPT-5-mini & 73.59 & 0.026 \\
    Qwen3-80B-A3B & 72.31 & -- \\
    \bottomrule
  \end{tabular*}
\end{table}

To verify the performance differences among various models serving as Judge-LLMs, we sampled 10 Wikipedia-DRA article pairs and manually annotated the win rate for each criterion, with total 390 criteria annotations. This allowed us to observe the consistency between different judge models and human judgments. We report the Pairwise Agreement Rate (PAR) for criterion-level evaluation across different models, with results shown in Table~\ref{tab:par_results}. Among the proprietary models, Gemini-2.5-pro demonstrated superior consistency. In open-source solutions, Qwen3-235B-A22B also shows strong consistency, maintaining competitive performance even compared to proprietary models.

\subsubsection{Wikipedia Leakage}

Since Wikipedia pages are openly accessible, data leakage where agent systems actively search for and read the original Wikipedia articles is a potential issue during the research process. Although our task prompts explicitly prohibited access to Wikipedia, some models failed to adhere to this instruction. To address this, during the evaluation of Cov. Wiki., we first filtered out statements that cited the original Wikipedia article as a source, performing retrieval only on the remaining statements. This ensured that the evaluation was not compromised by direct leakage from Wikipedia. We also calculated the statement-level leakage rate for different DRA systems, as shown in Table~\ref{tab:leakage}. These results provide insight into the models' ability to follow negative constraints. Notably, even with relatively high leakage rates (e.g., Perplexity Deep Research), performance scores remain suboptimal, indicating that mere access to Wikipedia does not guarantee the generation of unbiased and factually accurate articles. In contrast, LangChain (GPT-5) achieves high scores while maintaining an extremely low leakage rate, underscoring the robust capabilities of the system.

\begin{table}[!htbp]
  \centering
  \caption{\textbf{Wikipedia Leakage Rates across different DRAs.} The leakage rate indicates the proportion of statements directly citing the target Wikipedia page, reflecting model's adherence to the exclusion instruction.}
  \label{tab:leakage}
  \small
  \renewcommand{\arraystretch}{1.2}
  \begin{tabular*}{0.8\linewidth}{@{\extracolsep{\fill}}lc}
    \toprule
    \textbf{Models} & \textbf{Leakage Rate (\%)} \\
    \midrule
    Perplexity Deep Research & 33.77 \\
    Gemini-3-pro Deep Research & 23.35 \\
    Gemini-2.5-pro Deep Research & 14.74 \\
    Qwen-3-max Deep Research & 4.67 \\
    LangChain (GPT-4.1) & 1.13 \\
    OpenAI o3 Deep Research & 0.83 \\
    LangChain (GPT-5) & 0.09 \\
    Doubao Deep Research & 0.00 \\
    Grok Deep Search & 0.00 \\
    \bottomrule
  \end{tabular*}
\end{table}

\section{Conclusion}

In this study, we introduce \textbf{Wiki Live Challenge (WLC)}, a live benchmark that challenge the ability of Deep Research Agents to write Wikipedia-style articles, using Wikipedia Good Articles as human-expert references. WLC comprises 100 recent Good Articles spanning 15 categories, accompanied by a comprehensive evaluation framework, Wiki Eval. Wiki Eval combines a fine-grained writing assessment, constructing 39 criteria grounded in Wikipedia Good Article criteria, with a factual evaluation that measures both the coverage of Wikipedia facts and verifiability via cited references. Extensive experiments on a diverse set of deep research agents reveal a substantial gap between current agents and human-authored Wikipedia articles. We hope WLC will facilitate more reliable, fine-grained, and reproducible progress in the field of deep research agents.

\section*{Limitations}

Although the Wiki Live Challenge and Wiki Eval framework provide a comprehensive assessment of Deep Research Agents' capabilities, several limitations remain: (1) \textbf{Task Scale}: Due to constraints imposed by model knowledge cutoff dates, the number of collected Good Articles meeting our strict criteria is limited to the hundreds scale. Our benchmark prioritizes the quality and recency of Wikipedia articles over dataset size. (2) \textbf{System Opacity}: The lack of transparency in the citation mechanisms of certain proprietary systems, coupled with the potential inaccessibility of some cited web pages during evaluation, may impact the assessment of citation verifiability. Consequently, Reference Accuracy serves as an observational reference metric rather than a definitive measure of grounding.

\section*{Ethical Considerations}

\paragraph{Data Compliance.} Our benchmark dataset is derived entirely from Wikipedia, a publicly accessible knowledge base. The data collection process respects copyright policies and involves no personal privacy or non-public information.

\paragraph{Human Annotation and Compensation.} To validate our Wiki Eval evaluation framework and assess agent performance, we recruited 5 general annotators for collecting DRA result data and 5 PhD-level annotators for human evaluation. Participants were compensated at \$1 per article for collection and a rate of \$10 per hour for annotation. We obtained informed consent from all participants, and they were notified that the data would be used for research purposes only. The annotation tasks did not involve exposure to offensive or traumatic content.

\paragraph{Hallucinations and Misinformation.} DRAs have the potential to generate factually incorrect content or hallucinations, which poses significant risks in real-world applications. Our proposed benchmark specifically targets the evaluation of factual verifiability. By providing a rigorous standard for measuring hallucinations against expert-verified sources, our work contributes to the development of safer and more reliable AI systems.

\newpage
\bibliography{custom}
\bibliographystyle{acl_natbib}

\newpage
\appendix
\section{Details of Wiki Eval}\label{sec:appendixC}
Our criteria selection is grounded in the Wikipedia Good Article criteria, as shown in Figure~\ref{fig:ga_criteria}. We selected the first four dimensions (excluding stability and illustrations), focusing on the textual content. For the dimensions of \textit{Well-written}, \textit{Broad in its coverage}, and \textit{Neutral}, we delved into the specific guidelines for each, strictly adhering to Wikipedia's guidelines on \textit{lead sections}\footnote{\url{https://en.wikipedia.org/wiki/Wikipedia:Manual_of_Style/Lead_section}}, \textit{words to watch}\footnote{\url{https://en.wikipedia.org/wiki/Wikipedia:Manual_of_Style/Words_to_watch}}, \textit{verifiability}\footnote{\url{https://en.wikipedia.org/wiki/Wikipedia:Verifiability}}, \textit{summary style}\footnote{\url{https://en.wikipedia.org/wiki/Wikipedia:Summary_style}}, and \textit{neutral point of view}\footnote{\url{https://en.wikipedia.org/wiki/Wikipedia:Neutral_point_of_view}}, collecting a total of 39 fine-grained criteria to construct the \textbf{Wiki Writing} evaluation, which are listed in Table~\ref{tab:detailed_criteria_1}, Table~\ref{tab:detailed_criteria_2}, and Table~\ref{tab:detailed_criteria_3}. For the \textit{Verifiable} dimension, we constructed the \textbf{Wiki Fact}, an evaluation metric that encompasses both the factual accuracy of the generated article relative to Wikipedia and its consistency with cited references.

\begin{figure*}[t]
  \centering
  \begin{promptbox}[The six Good Article criteria]
    \textbf{1. Well-written:} \\
    (a) the prose is clear, concise, and understandable to an appropriately broad audience; spelling and grammar are correct; \\
    (b) it complies with the Manual of Style guidelines for lead sections, layout, words to watch, fiction, and list incorporation. \\
    \textbf{2. Verifiable with no original research:} \\
    (a) it contains a list of all references (sources of information), presented in accordance with the layout style guideline; \\
    (b) reliable sources are cited inline. All content that could reasonably be challenged, except for plot summaries and that which summarizes cited content elsewhere in the article, must be cited no later than the end of the paragraph (or line if the content is not in prose); \\
    (c) it contains no original research; and \\
    (d) it contains no copyright violations or plagiarism. \\
    \textbf{3. Broad in its coverage:} \\
    (a) it addresses the main aspects of the topic; and \\
    (b) it stays focused on the topic without going into unnecessary detail (see summary style). \\
    \textbf{4. Neutral:} it represents viewpoints fairly and without editorial bias, giving due weight to each. \\
    \textcolor{gray!60}{\textbf{5. Stable:} it does not change significantly from day to day because of an ongoing edit war or content dispute. (Not applicable for offline evaluation)} \\
    \textcolor{gray!60}{\textbf{6. Illustrated, if possible, by media such as images, video, or audio:} (Not applicable for text-only evaluation)} \\
    \textcolor{gray!60}{(a) media are tagged with their copyright statuses, and valid non-free use rationales are provided for non-free content; and} \\
    \textcolor{gray!60}{(b) media are relevant to the topic, and have suitable captions.}
  \end{promptbox}
  \caption{\textbf{The six Wikipedia Good Article criteria.} These criteria serve as the foundation for our Wiki Eval framework.}
  \label{fig:ga_criteria}
\end{figure*}

\begin{table*}[p]
  \centering
  \small
  \renewcommand{\arraystretch}{1.3} 
  \begin{tabular}{|p{0.97\textwidth}|}
    \toprule
    \textbf{Category 1: Well-written (Part 1)} \\
    \midrule
    \textbf{Clear \& Concise:} \\
    1. Articles should be written in encyclopedic style, which differs from the technically dense style found in scholarly writing aimed at specialists. \\
    2. Articles should address the topic without twisting the truth or telling "lies-to-children", but should also minimize (unexplained) jargon and not take prior knowledge for granted. \\
    3. Articles should be self-contained as much as possible, rather than relying on excessive links to explain unfamiliar concepts. \\
    \textbf{Lead section:} \\
    1. An article's content should begin with an introductory lead section – a concise summary of the article – which is never divided into sections. The remainder of the article is typically divided into sections. \\
    2. The lead section gives the basics in a nutshell, introduces the article, and cultivates interest in reading on—though not by teasing the reader or hinting at what follows. \\
    3. The lead section should be written in a clear, accessible style with a neutral point of view. \\
    4. The lead should stand on its own as a concise overview of the article's topic. It should identify the topic, establish context, explain why the topic is notable, and summarize the most important points, including any prominent controversies. The notability of the article's subject is usually established in the first few sentences. \\
    5. As in the body of the article itself, the emphasis given to material in the lead should roughly reflect its importance to the topic, according to reliable, published sources. Apart from basic facts, significant information should not appear in the lead if it is not covered in the remainder of the article. \\
    6. A lead section should be carefully sourced as appropriate, although it is common for citations to appear in the body and not the lead. \\
    7. Significant information should not appear in the lead, apart from basic facts, if it is not covered in the remainder of the article, although not everything in the lead must be repeated in the body of the text. Exceptions include specific facts such as quotations, examples, birth dates, taxonomic names and typological classifications, case numbers, and titles. \\
    \bottomrule
  \end{tabular}
  \caption{Detailed fine-grained criteria for \textbf{Well-written} (Part 1).}
  \label{tab:detailed_criteria_1}
\end{table*}

\begin{table*}[p]
  \centering
  \small
  \renewcommand{\arraystretch}{1.3}
  \begin{tabular}{|p{0.97\textwidth}|}
    \toprule
    \textbf{Category 1: Well-written (Part 2)} \\
    \midrule
    \textbf{Words to Watch:} \\
    1. Puffery: Words such as these are often used without attribution to promote the subject of an article, while neither imparting nor plainly summarizing verifiable information. They are known as "peacock terms" by Wikipedia contributors. Instead of making subjective proclamations about a subject's importance, use facts and attribution to demonstrate it. Words to Watch: legendary, best, great, greatest, acclaimed, iconic, visionary, outstanding, leading, celebrated, popular, award-winning, landmark, cutting-edge, innovative, revolutionary, extraordinary, brilliant, hit, famous, renowned, remarkable, prestigious, world-class, respected, notable, virtuoso, honorable, awesome, unique, pioneering, phenomenal, prominent. \\
    2. Contentious labels: Value-laden labels – such as calling an organization a cult, an individual a racist, sexist, terrorist, or freedom fighter, or a sexual practice a perversion – may express contentious opinion and are best avoided unless widely used by reliable sources to describe the subject, in which case use in-text attribution. Avoid myth in its informal sense, and establish the scholarly context for any formal use of the term. Words to Watch: cult, racist, perverted, sexist, homophobic, transphobic, misogynistic, sect, fundamentalist, heretic, extremist, denialist, terrorist, freedom fighter, bigot, myth, neo-Nazi, -gate, pseudo-, controversial. \\
    3. Unsupported attributions: These words may disguise a biased view. Claims about what people say, think, feel, or believe, and what has been shown, demonstrated, or proved should be clearly attributed. Words to Watch: some people say, many people remember, many scholars state, it is believed/regarded/considered, many are of the opinion, most feel, experts declare, it is often reported, it is widely thought, research has shown, science says, scientists claim, it is often said, officially, is widely regarded as, X has been described as Y. \\
    4. Expressions of doubt: When these are used, ensure that the source of the accusation is clear. Words to Watch: supposed, apparent, purported, alleged, accused, so-called ... Also, scare-quoting: a Yale "report"; undue emphasis: "... a Baptist church". \\
    5. Editorializing: These words should usually be avoided so as to maintain an impartial tone. Words to Watch: notably, it should be noted, arguably, interestingly, essentially, utterly, actually, only, clearly, absolutely, of course, without a doubt, indeed, happily, sadly, tragically, aptly, fortunately, unfortunately, untimely. \\
    6. Synonyms for said: In some types of writing, repeated use of said is considered tedious, and writers are encouraged to employ synonyms. On Wikipedia, it is more important to avoid language that makes undue implications. Words to Watch: reveal, point out, clarify, expose, explain, find, note, observe, insist, speculate, surmise, claim, assert, admit, confess, deny, confirm ... \\
    7. Euphemisms: Euphemisms should generally be avoided in favor of more neutral and precise terms. Words to Watch: passed away, gave her life, eternal rest, make love, an issue with, collateral damage, differently abled. \\
    8. Clichés and idioms: Clichés and idioms should generally be avoided in favor of direct, literal expressions. Words to watch: lion's share, tip of the iceberg, white elephant, gild the lily, take the plunge, ace up the sleeve, bird in the hand, twist of fate, at the end of the day. \\
    9. Relative time references: Prefer specific statements to general ones. Words to watch: recently, lately, currently, today, presently, to date, X years ago, formerly, in the past, traditionally, this/last/next (year/month/winter/spring/summer/fall/autumn), yesterday, tomorrow, in the future, now, to this day, soon, since. \\
    10. Unspecified places or events: Prefer specific statements to general ones. Words to watch: this country, here, there, somewhere, sometimes, often, occasionally, somehow. \\
    11. Survived by: Phrasing such as "Smith died in 1982, survived by her husband Jack and two sons" should be avoided; this information can be made more complete and spread out through the article. Words to watch: is/was survived by, [Name]'s survivors include. \\
    \bottomrule
  \end{tabular}
  \caption{Detailed fine-grained criteria for \textbf{Well-written} (Part 2).}
  \label{tab:detailed_criteria_2}
\end{table*}

\begin{table*}[p]
  \centering
  \small
  \renewcommand{\arraystretch}{1.3}
  \begin{tabular}{|p{0.97\textwidth}|}
    \toprule
    \textbf{Category 2: Broad in its coverage} \\
    \midrule
    1. It addresses the main aspects of the topic. \\
    2. The lead, ideally the introductory sentence or at least introductory paragraph, of an article, should make clear what the scope of the article is. \\
    3. All material that is notable, referenced, and that a reader would be likely to agree matches the specified scope must be covered (at least in a summarized fashion). \\
    4. What reliable sources say about material that is out of scope for the decided-upon subject is largely irrelevant to that article and can be removed. \\
    5. It stays focused on the topic without going into unnecessary detail. \\
    6. The lead contains a quick summary of the topic's most important points. \\
    7. Each major subtopic is detailed in its own section of the article. \\
    8. It contains no irrelevant (nor only loosely relevant) information. \\
    \midrule
    \textbf{Category 3: Neutral} \\
    \midrule
    1. It represents viewpoints fairly and without editorial bias, giving due weight to each. \\
    2. Avoid stating opinions as facts. Usually, articles will contain information about the significant opinions that have been expressed about their subjects. However, these opinions should not be stated in Wikipedia's voice. Rather, they should be attributed in the text to particular sources, or where justified, described as widespread views, etc. For example, an article should not state that genocide is an evil action but may state that genocide has been described by John So-and-so as the epitome of human evil. \\
    3. Avoid stating seriously contested assertions as facts. If different reliable sources make conflicting assertions about a matter, treat these assertions as opinions rather than facts, and do not present them as direct statements. \\
    4. Avoid stating facts as opinions. Uncontested and uncontroversial factual assertions made by reliable sources should normally be directly stated in Wikipedia's voice, for example the sky is blue not [name of source] believes the sky is blue. Unless a topic specifically deals with a disagreement over otherwise uncontested information, there is no need for specific attribution for the assertion, although it is helpful to add a reference link to the source in support of verifiability. Further, the passage should not be worded in any way that makes it appear to be contested. \\
    5. Prefer nonjudgmental language. A neutral point of view neither sympathizes with nor disparages its subject (or what reliable sources say about the subject), although this must sometimes be balanced against clarity. Present opinions and conflicting findings in a disinterested tone. Do not editorialize. When editorial bias towards one particular point of view can be detected the article needs to be fixed. The only bias that should be evident is the bias attributed to the source. \\
    6. Indicate the relative prominence of opposing views. Ensure that the reporting of different views on a subject adequately reflects the relative levels of support for those views and that it does not give a false impression of parity, or give undue weight to a particular view. For example, to state that According to Simon Wiesenthal, the Holocaust was a program of extermination of the Jewish people in Germany, but David Irving disputes this analysis would be to give apparent parity between the supermajority view and a tiny minority view by assigning each to a single activist in the field. \\
    7. An article should not give undue weight to minor aspects of its subject but should strive to treat each aspect with a weight proportional to its treatment in the body of reliable, published material on the subject. \\
    8. Any inclusion of fringe or pseudoscientific views should not give them undue weight. The fringe or pseudoscientific view should be clearly described as such. An explanation of how experts in the relevant field have reacted to such views should be prominently included. \\
    9. In the case of beliefs and practices, Wikipedia content should not only encompass what motivates individuals who hold these beliefs and practices but also account for how such beliefs and practices developed. Wikipedia articles on history and religion draw from religion's sacred texts as primary sources and modern archaeological, historical, and scientific works as secondary and tertiary sources. \\
    10. Article titles that combine alternative names are discouraged. For example, names such as "Derry/Londonderry", "Aluminium/Aluminum", and "Flat Earth (Round Earth)" should not be used. Instead, alternative names should be given their due prominence within the article itself, and redirects created as appropriate. \\
    \bottomrule
  \end{tabular}
  \caption{Detailed fine-grained criteria for \textbf{Broad in its coverage} and \textbf{Neutral}.}
  \label{tab:detailed_criteria_3}
\end{table*}

\section{Detailed Evaluation Settings}\label{sec:appendixA}
\subsection{Implementation Details}\label{sec:appendixA.1}
\paragraph{Judge LLM for Wiki Writing}
Based on the agreement with human evaluation results, we selected Gemini-2.5-pro, which demonstrated the best performance, as the Judge-LLM for assessing the win rate. During evaluation, we input the Wikipedia original article, the generated article, and the criteria belonging to the same category as a single prompt into the model for batch evaluation. The evaluation prompt is shown in Figure~\ref{fig:criteria_prompt}.

\begin{figure*}[t]
  \centering
  \begin{promptbox}[System Prompt]
    You are a strict, meticulous, and objective Wikipedia article evaluation expert. You excel at using specific criteria to compare two Wikipedia-style articles on the same topic, and for each criteria you MUST decide a clear winner - either Article 1 wins or Article 2 wins.
  \end{promptbox}
  \begin{promptbox}[User Prompt]
    \textbf{Articles to Evaluate} \\
    $<$article\_1$>$ \\
    \{article1\_text\} \\
    $<$/article\_1$>$ \\
    $<$article\_2$>$ \\
    \{article2\_text\} \\
    $<$/article\_2$>$ \\
    \textbf{Criterion Category and Criteria} \\
    We focus on: \{category.name\} \\
    \{category.description\} \\
    Under this category there are \{k\} criteria: \\
    \{criteria\_list\} \\
    Compare which article better satisfies each criteria criterion. \\
    Output your response in strict JSON format with exactly \{k\} results (one per criteria): \\
    \{ \\
      "results": [ \\
        \{ "criteria\_index": 1, "reason": "Brief explanation for criteria 1", "winner": 1 \}, \\
        \{ "criteria\_index": 2, "reason": "Brief explanation for criteria 2", "winner": 2 \} \\
      ] \\
    \} \\
    Note: winner value should be: \\
    - 1 = Article 1 wins \\
    - 2 = Article 2 wins
  \end{promptbox}
  \caption{\textbf{Prompt used for Wiki Writing Evaluation.} The model is tasked with comparing two articles based on specific criteria and determining a winner.}
  \label{fig:criteria_prompt}
\end{figure*}

\paragraph{Judge LLM for Wiki Fact}
We employ Gemini-2.5-flash as both the fact extraction model and the fact-checking model. The prompts for fact extraction and fact checking are presented in Figure~\ref{fig:extract_prompt} and Figure~\ref{fig:check_prompt}, respectively.

\begin{figure*}[t]
  \centering
  \begin{promptbox}[Fact Extraction Prompt]
    You will be provided with an article. The body of the article contains text with citations to references. \\
    Citations in the main text may appear in the following forms: ...[1], ... .1, ...[(1)](https://...) \\
    \textbf{Your Task:} \\
    Please extract \textbf{all} informative statements (facts) from the main text. Each statement \textbf{must} be associated with its corresponding citation number(s). Return the result as a list of (fact, ref\_idx) pairs. \\
    \textbf{Citation Assignment Rule (Critical):} \\
    For each statement, find the \textbf{nearest citation(s) that immediately follow it} in the text. This is the citation that covers this statement. \\
    - If a statement is followed directly by one or more citations (e.g., [1] or [1][2][3]), those are the citations for that statement. \\
    - If a statement has no citation immediately after it, look forward in the text to find the \textbf{next citation(s)} that appear. All statements between the previous citation and this next citation share this next citation as their source. \\
    - In other words: citations at the end of a sentence or paragraph typically cover all the preceding uncited statements up to the last citation marker. \\
    \textbf{Extraction Rules:} \\
    1. \textbf{Completeness \& Context (Crucial):} Every extracted fact must be a complete, self-contained statement. Do not extract fragmented phrases. \\
    2. \textbf{Handling Citations:} Each statement must have a citation. Use the nearest following citation(s) as described above. \\
    3. \textbf{Formatting:} Return a JSON list. Ensure Chinese quotation marks in the fact are preserved. \\
    \textbf{Output Format Example:} \\
    Given text: "Li Qiang constructed a socioeconomic status index...[15] It has been validated...[3][7][12]" \\
    Output: [ \{ "fact": "...", "ref\_idx": "15" \}, \{ "fact": "...", "ref\_idx": ["3", "7", "12"] \} ] \\
    Here is the main text of the article: \\
    \{article\_text\} \\
    Please begin the extraction now. Output only the JSON list directly, without any chitchat or explanations.
  \end{promptbox}
  \caption{\textbf{Prompt used for Fact Extraction.} The model extracts atomic facts and their associated citation indices from the text.}
  \label{fig:extract_prompt}
\end{figure*}

\begin{figure*}[t]
  \centering
  \begin{promptbox}[System Prompt]
    You are a professional data annotator. Given a sentence and a paragraph composed of several gold factual support sentences, your task is to verify whether the sentence is consistent with the facts. Matching any single support sentence is enough for consistency. \\
    Decision policy: \\
    consistent: At least one support sentence directly entails or clearly confirms the statement. \\
    inconsistent: At least one support sentence clearly contradicts the statement. \\
    not\_support: No support sentence clearly confirms or contradicts the statement (insufficient evidence).
  \end{promptbox}
  \begin{promptbox}[User Prompt]
    Below are the sentence to judge and the factual paragraph: \\
    $<$judge\_sentence$>$ \\
    \{gen\_sentence\} \\
    $<$/judge\_sentence$>$ \\
    $<$factual\_paragraph$>$ \\
    \{gt\_sentence\} \\
    $<$/factual\_paragraph$>$ \\
    Output Format (strict JSON only): \\
    \{ \\
    "reason": "...", \\
    "verdict": "consistent | inconsistent | not\_support" \\
    \}
  \end{promptbox}
  \caption{\textbf{Prompt used for Fact Checking.} The model verifies the consistency of a generated statement against a gold factual paragraph.}
  \label{fig:check_prompt}
\end{figure*}
\subsection{Details of Evaluated DRAs}\label{sec:appendixA.2}
\subsubsection{Data Collection Process}
\paragraph{DRAs with Web Services} Data for Gemini-2.5-pro Deep Research, OpenAI o3 Deep Research, Perplexity Deep Research, Grok Deep Search, Qwen-3-max Deep Research, and Doubao Deep Research were collected via their respective web services. During collection, some DRA systems required secondary user interaction to confirm the research direction; in these cases, all human annotators are instructed to use a unified prompt instruction: \textit{All content requires you to research and follow the criteria I provide. Always exclude the search or reading of Wikipedia pages.}

\paragraph{Locally Deployed DRAs} Data for LangChain Open Deep Research, Deep Researcher, and Tongyi Deep Research were collected in a local environment. For LangChain Open Deep Research, we followed its open-source framework settings, using Tavily\footnote{\url{https://www.tavily.com/}} as the search engine, and employed GPT-4.1 and GPT-5 as the backbone models for article generation. For Deep Researcher, we deployed its model on a single H20 GPU and used the crawling tools provided in its open-source repository for article generation. For Tongyi Deep Research, we deployed its model on a single H20 GPU, using Serper API\footnote{\url{https://serper.dev/}} as the search engine and Jina\footnote{\url{https://jina.ai/}} as the web crawling tool.

\paragraph{DRAs via API Services} The recently released Gemini Deep Research Agent by Google~\footnote{\url{https://ai.google.dev/gemini-api/docs/deep-research}}, powered by Gemini-3-Pro, is capable of navigating complex information environments using web search to generate detailed reports with citations. We generated articles by invoking the Gemini Deep Research Agent API service.

\subsubsection{Data Collection Costs}
We report the costs incurred in collecting 100 articles generated by each DRA system. This encompasses both the compensation for human annotators and the operational costs of the DRA systems. Human annotators were paid \$1 per article collection for their time. The specific generation costs for each DRA system are detailed in Table~\ref{tab:collection_costs}.

\begin{table*}[t]
  \centering
  \small
  \setlength{\tabcolsep}{12pt}
  \renewcommand{\arraystretch}{1.2}
  \begin{tabular}{lccc}
    \toprule
    \textbf{Models} & \textbf{Account/API Cost (\$)} & \textbf{Human Cost (\$)} & \textbf{Total (\$)} \\
    \midrule
    \multicolumn{4}{c}{\textit{Open-Source Agent Framework}} \\
    \midrule
    Deep Researcher & -- & -- & -- \\
    Tongyi Deep Research & -- & -- & -- \\
    Langchain Open Deep Research (GPT-4.1) & 10.7 & -- & 10.7 \\
    Langchain Open Deep Research (GPT-5) & 460.9 & -- & 460.9 \\
    \midrule
    \multicolumn{4}{c}{\textit{Proprietary Agent Framework}} \\
    \midrule
    Doubao Deep Research & -- & 100 & 100 \\
    Qwen-3-max Deep Research & -- & 100 & 100 \\
    Perplexity Deep Research & 40 & 100 & 140 \\
    Grok Deep Search & -- & 100 & 100 \\
    OpenAI o3 Deep Research & 200 & 100 & 300 \\
    Gemini-2.5-pro Deep Research & 100 & 100 & 200 \\
    Gemini-3-pro Deep Research & 157.3 & -- & 157.3 \\
    \bottomrule
  \end{tabular}
  \caption{\textbf{Cost of collecting 100 articles for each DRA system.} Account/API Cost refers to fees for model access or subscription. Human Cost refers to compensation for annotators collecting data via web services (\$1/article). ``--'' indicates negligible costs (e.g., free access or automated collection).}
  \label{tab:collection_costs}
\end{table*}

\section{Detailed Analysis Across Categories}\label{sec:appendixB}
\subsection{Difficulty Analysis}\label{sec:appendixB.1}
Table~\ref{tab:category_difficulty} presents the detailed difficulty ranking of different Wikipedia categories. The difficulty is measured by the average win rate in Wiki Writing Criteria of all DRA systems in each category. We also list the average article length, statement count, external link count, and total page views for each category.

\begin{table*}[h]
  \centering
  \small
  \setlength{\tabcolsep}{8pt}
  \renewcommand{\arraystretch}{1.2}
  \begin{tabular}{clccccc}
    \toprule
    \textbf{Rank} & \textbf{Category} & \textbf{Difficulty (\%)} & \textbf{Avg. Bytes} & \textbf{Avg. Stmts} & \textbf{Avg. Links} & \textbf{Avg. Views} \\
    \midrule
    1 & History & 16.6 & 42,882 & 175.8 & 45.5 & 16,927 \\
    2 & Mathematics & 18.9 & 29,445 & 85.0 & 47.0 & 5,648 \\
    3 & Language and literature & 19.6 & 55,728 & 177.8 & 94.8 & 4,786 \\
    4 & Art and architecture & 22.0 & 49,681 & 145.7 & 102.9 & 10,051 \\
    5 & Engineering and technology & 22.2 & 57,869 & 169.1 & 94.0 & 6,721 \\
    6 & Agriculture, food and drink & 25.2 & 37,317 & 70.0 & 111.0 & 5,756 \\
    7 & Media and drama & 27.0 & 39,891 & 126.1 & 70.1 & 179,130 \\
    8 & Music & 28.4 & 36,845 & 111.9 & 84.9 & 62,728 \\
    9 & Video games & 29.0 & 32,491 & 89.7 & 56.3 & 7,907 \\
    10 & Sports and recreation & 30.2 & 41,862 & 151.3 & 73.1 & 18,342 \\
    11 & Warfare & 32.0 & 48,352 & 223.7 & 37.8 & 8,757 \\
    12 & Social sciences and society & 32.6 & 51,052 & 132.5 & 85.8 & 12,414 \\
    13 & Geography and places & 38.9 & 46,078 & 162.0 & 94.0 & 1,024 \\
    14 & Natural sciences & 41.8 & 50,660 & 186.6 & 90.1 & 47,984 \\
    15 & Philosophy and religion & 56.2 & 39,287 & 142.0 & 69.0 & 154,968 \\
    \bottomrule
  \end{tabular}
  \caption{\textbf{Difficulty ranking and statistical features of different Wikipedia categories.} Difficulty is defined as the average win rate in Wiki Writing Criteria of all DRA systems in that category.}
  \label{tab:category_difficulty}
\end{table*}

\subsection{Robustness Analysis}\label{sec:appendixB.2}
To verify the robustness of our evaluation across different Wikipedia categories, we formulate the null hypothesis ($H_0$) that there is no significant difference in the relative performance of a DRA system when evaluated across different categories. The relative performance is calculated as the deviation of the system's score from the category mean to control for inherent category difficulty.

We conducted an ANOVA test for each system to test this hypothesis. The detailed results, including $p$-values and effect sizes ($\eta^2$), are presented in Table~\ref{tab:anova_results}.

\begin{table*}[t]
  \centering
  \small
  \setlength{\tabcolsep}{12pt}
  \renewcommand{\arraystretch}{1.2}
  \begin{tabular}{lcc}
    \toprule
    \textbf{System} & \textbf{$p$-value} & \textbf{$\eta^2$} \\
    \midrule
    Doubao Deep Research & 0.9781 & 0.040 \\
    Grok Deep Search & 0.9065 & 0.059 \\
    Gemini-2.5-pro Deep Research & 0.8831 & 0.063 \\
    OpenAI o3 Deep Research & 0.8669 & 0.066 \\
    Tongyi DeepResearch & 0.6097 & 0.097 \\
    Qwen-3-max Deep Research & 0.5770 & 0.101 \\
    LangChain (GPT-4.1) & 0.5702 & 0.101 \\
    LangChain (GPT-5) & 0.5345 & 0.105 \\
    Gemini-3-pro Deep Research & 0.3200 & 0.132 \\
    Perplexity Deep Research & 0.1615 & 0.161 \\
    Deep Researcher & 1.11e-11 & 0.571 \\
    \bottomrule
  \end{tabular}
  \caption{\textbf{ANOVA test results for cross-category robustness.} A high $p$-value ($>0.05$) and low effect size ($\eta^2$) indicate strong evidence for the null hypothesis, supporting the robustness of the evaluation across categories.}
  \label{tab:anova_results}
\end{table*}
\section{Human Annotation}\label{sec:appendixD}
We recruited five PhD-level annotators and tasked them with evaluating randomly sampled pairs of Wikipedia articles and model-generated articles based on each criterion. All article data underwent preprocessing to rigorously remove references and inline citation tags, and was presented in Markdown format. The order of articles was randomized to ensure annotators evaluated solely based on writing quality. We strictly instructed annotators to thoroughly read the articles under comparison. For each criterion, in addition to determining the winner, they were required to provide a rationale for their decision. Annotators were compensated at a rate of \$10 per hour.
\end{document}